\documentclass[10pt,twocolumn,letterpaper]{article}

\usepackage{cvpr}
\usepackage{times}
\usepackage{epsfig}
\usepackage{graphicx}
\usepackage{amsmath}
\usepackage{amssymb}
\usepackage{bbm}
\usepackage{marvosym}
\usepackage{subfigure}
\usepackage{algorithm}
\usepackage{algorithmic}
\usepackage{booktabs}
\usepackage{multirow}
\usepackage[table]{xcolor}
\usepackage{colortbl}
\usepackage{flushend}

\DeclareMathOperator*{\argmax}{arg\,max}
\usepackage[pagebackref=true,breaklinks=true,letterpaper=true,colorlinks,bookmarks=false]{hyperref}

\cvprfinalcopy 

\ifcvprfinal\fi
\begin{document}

\title{Newtonian Image Understanding:\\ Unfolding the Dynamics of Objects in Static Images}
\author{Roozbeh Mottaghi$^\dagger$ \qquad Hessam Bagherinezhad$^\ddagger$
\qquad Mohammad Rastegari$^\dagger$
\qquad Ali Farhadi$^{\dagger\ddagger}$\\
$^\dagger$Allen Institute for Artificial Intelligence (AI2)\\
$^\ddagger$University of Washington
}

\maketitle

\begin{abstract}
In this paper, we study the challenging problem of predicting the dynamics of objects in static images. Given a query object in an image, our goal is to provide a physical understanding of the object in terms of the forces acting upon it and its long term motion as response to those forces. Direct and explicit estimation of the forces and the motion of objects from a single image is extremely challenging. We define intermediate physical abstractions called Newtonian scenarios and introduce Newtonian Neural Network ($N^3$) that learns to map a single image to a state in a Newtonian scenario. Our experimental evaluations show that our method can reliably predict dynamics of a query object from a single image. In addition, our approach can provide physical reasoning that supports the predicted dynamics in terms of velocity and force vectors. To spur research in this direction we compiled Visual Newtonian Dynamics (VIND) dataset that includes 6806 videos aligned with Newtonian scenarios represented using game engines, and 4516 still images with their ground truth dynamics.
\end{abstract}


\section{Introduction}
A key capability in human perception is the ability to proactively predict what happens next in a scene \cite{bar09}. Humans reliably use these predictions for planning their actions, making everyday decisions , and even correcting visual interpretations~\cite{hawkins04}. Examples include predictions involved in passing a busy street, catching a frisbee, or hitting a tennis ball with a racket. Performing these tasks require  a  rich understanding of the dynamics of  objects moving in a scene. For example, hitting a tennis ball with a racket requires knowing the dynamics of the ball, when it hits the ground, how it bounces back from the ground, and what form of motion it follows. 

\begin{figure}
\centering
  \includegraphics[width=18pc]{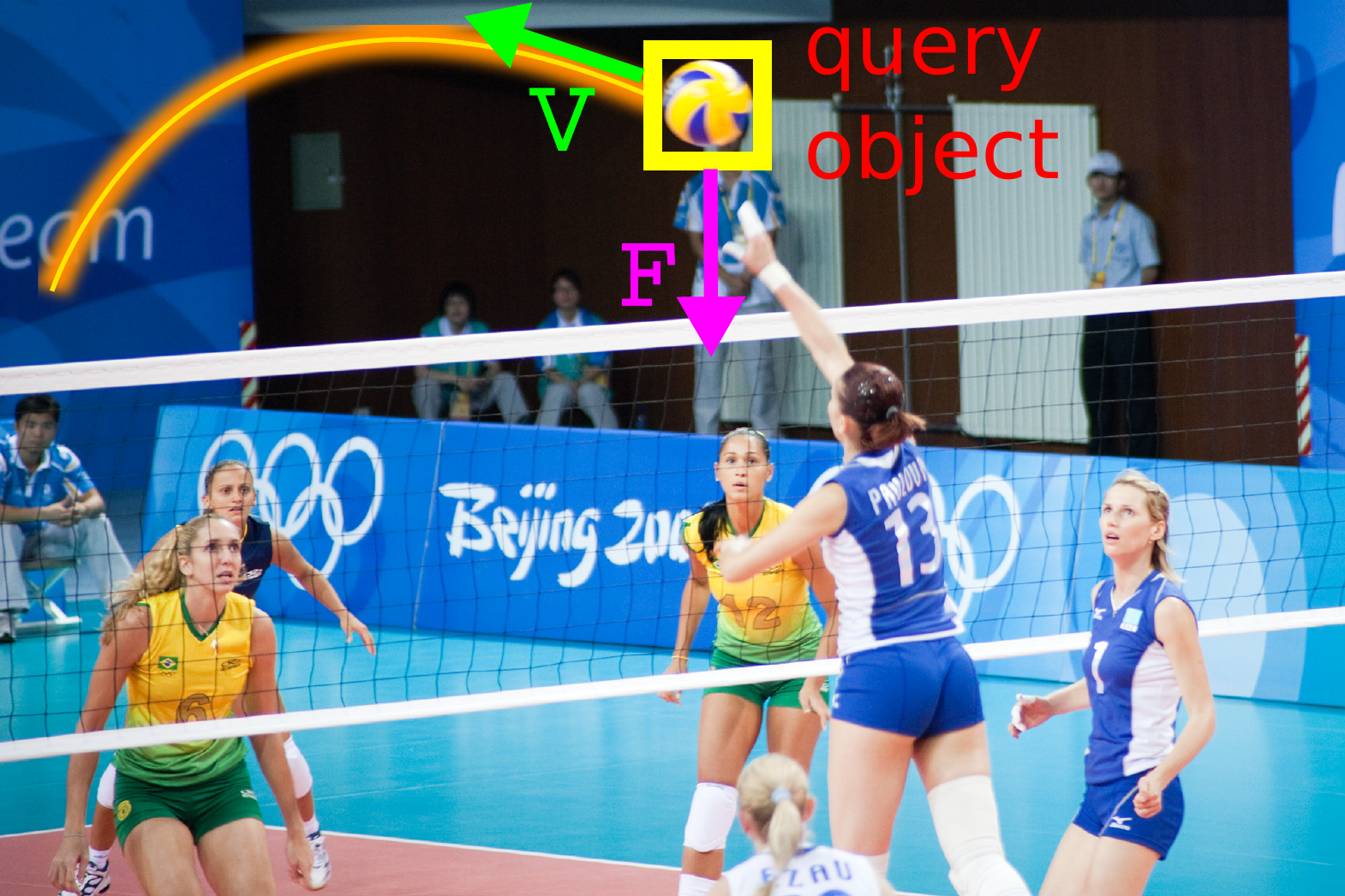}
\caption{Given a static image, our goal is to infer the dynamics of a query object (forces that are acting upon the object and the expected motion of the object as a response to those forces). In this paper, we show an algorithm that learns to map an image to a state in a physical abstraction called a Newtonian scenario. Our method provides a rich physical understanding of an object in an image that allows prediction of long term motion of the object and  reasoning about the direction of net force and velocity vectors. }
\label{fig:teaser}
\end{figure}

Rich physical understanding of human perception even allows predictions of dynamics on only a single image. Most people, for example, can reliably predict the dynamics of the volleyball shown in Figure~\ref{fig:teaser}. Theories in perception and cognition attribute this capability, among many explanations, to previous experience~\cite{cheung12} and existence of an underlying physical abstraction~\cite{hamrick11}. 

\begin{figure*}
\centering
  \includegraphics[width=35pc]{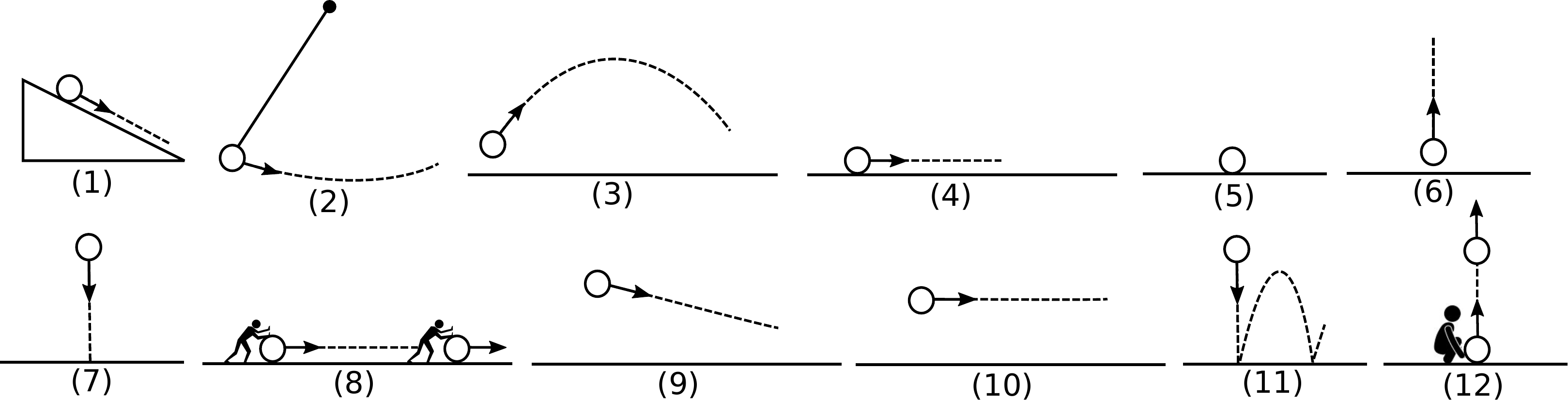}
\caption{\textbf{Newtonian Scenarios} are defined according to different physical quantities: direction of motion, forces, etc. We use 12 scenarios that are depicted here. The circle represents the object, and the arrow shows the direction of its motion.}
\label{fig:motion}
\end{figure*} 

In this paper, we address the problem of physical understanding of objects in images in terms of the forces actioning upon them and their long term motions as their responses to those forces. Our goal is to unfold the dynamics of objects in still images. Figure~\ref{fig:teaser} shows an example of a long term motion predicted by our approach along with the physical reasoning that supports the predicted dynamics.  

Motion of objects and its relations to various physical quantities (mass, friction, external forces, geometry, etc.) has been extensively studied in Mechanics. In schools, classical mechanics is taught using basic Newtonian scenarios that explain a large number of simple motions in real world: inclined surfaces, falling, swinging, external forces, projectiles, etc. To infer the dynamics of an object, students need to figure out the Newtonian scenario that explains the situation, find the physical quantities that contribute to the motion, and then plug them into the corresponding equations that relate contributing physical quantities to the motion.

Estimating physical quantities from an image is an extremely challenging problem. For example, computer vision literature does not provide a reliable solution to direct estimation of mass, friction, the angle of an inclined plane, etc. from an image. Instead of direct estimation of the physical quantities from images,  we formulate the problem of physical understanding as a mapping from an image to a physical abstraction. We follow the same principles of classical Mechanics and use Newtonian scenarios as our physical abstraction. These scenarios are depicted in Figure~\ref{fig:motion}. We chose to learn this mapping in the visual space and thus render the Newtonian scenarios using game engines. 

Mapping a single image to a state in a Newtonian scenario allows us to borrow the rich Newtonian interpretation offered by game engines. This enables predicting the long term motion of the object along with rich physical reasoning that supports the predicted motion in terms of velocity and force vectors\footnote{Throughout this paper we refer to force and velocity vector as  normalized unit vectors that show the direction of force or velocity.}. Learning such a mapping requires reasoning about subtle visual and contextual cues, and common knowledge of motion.  For example, to predict the expected motion of the ball in Figure~\ref{fig:teaser} one needs to rely on previous experience, visual cues (subtle hand posture of the player on the net, the line of sight of other players, their pose, scene configuration), and the knowledge about how objects move in a volleyball scene. To perform this mapping, we adopt a data driven approach and introduce Newtonian Neural Networks ($N^3$) that learns the complex interplay between visual cues and motions of objects.

To facilitate research in this challenging  direction, we compiled \textit{VIND}, VIsual Newtonian Dynamics dataset, that contains 6806 videos, with the corresponding game engine videos for training and 4516 still images with the predicted motions for testing. 

Our experimental evaluations show promising results in Newtonian understanding of objects in images and enable prediction of long-term motions of objects backed by abstract Newtonian explanations of the predicted dynamics. This allows us to unfold the dynamics of moving objects in static images. Our experimental evaluations also show the benefits of using an intermediate physical abstraction compared to competitive baselines that make direct predictions of the motion.

\section{Related Work}
\textbf{Cognitive studies:} Recent studies in computational cognitive science show that humans approximate the principles of Newtonian dynamics and simulate the future states of the world using these principles \cite{hamrick11, battaglia13}. Our use of Newtonian scenarios as an intermediate representation is inspired by these studies.
 
\textbf{Motion prediction:} The problem of predicting future movements and trajectories has been tackled from different perspectives. Data-driven approaches have been proposed in \cite{yuen10, liu11} to predict motion field in a single image.  Future trajectories of people are inferred in \cite{kitani12}. \cite{walker14} proposed to infer the most likely path for objects. In contrast, our method focuses on the physics of the motion and estimates a 3D long-term motion for objects. There are recent methods that address prediction of optical flow in static images \cite{pintea14,walker15}. Flow does not carry semantics and represents very short-term motions in 2D whereas our method can infer long term 3D motions using force and velocity information. Physic-based human motion modeling was studied by \cite{brubaker2009,brubaker2008,brubaker2007,vondrak2008}. They employed human movement dynamics to predict future pose of humans. In contrast, we estimate the dynamics of objects. 

\begin{figure*}
\centering
  \includegraphics[width=33pc]{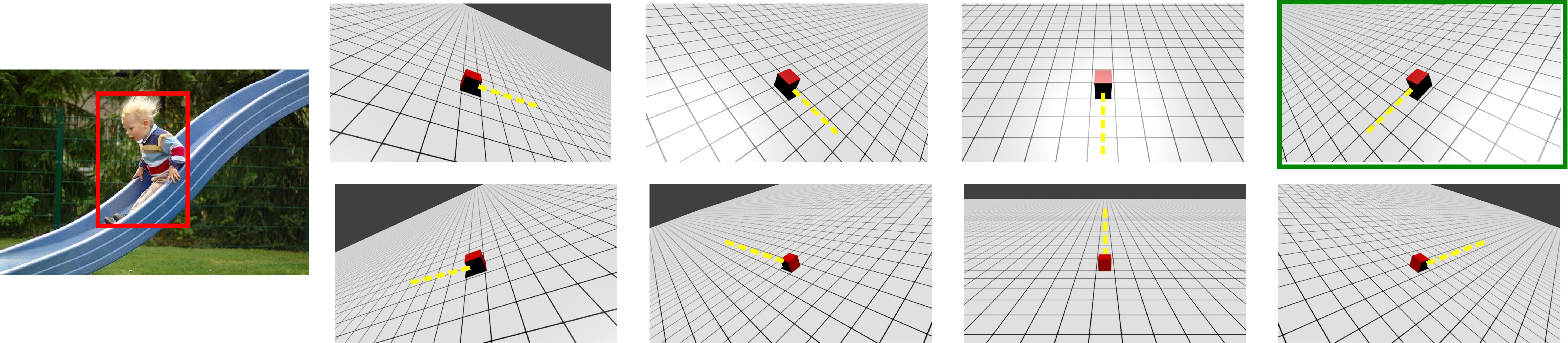}
\caption{\textbf{Viewpoint annotation.} We ask the annotators to choose the game engine video (among 8 different views of the Newtonian scenario) that best describes the view of the object in the image. The object in the game engine video is shown in red, and its direction of movement is shown in yellow. The video with a green border is the selected viewpoint. These videos correspond to Newtonian scenario (1).}
\label{fig:annimages}
\end{figure*}

\textbf{Scene understanding:} Reasoning about the stability of a scene has been addressed in \cite{jia13} that use physical constraints to reason about the stability of objects that are modeled by 3D volumes. Our work is different in that we reason about the dynamics of stable and moving objects. The approach of \cite{zheng14} computes the probability that an object falls based on inferring disturbances caused naturally or by human actions. In contrast, we do not explicitly encode physics equations and we rely on images and direct perception. The early work of Mann \etal \cite{mann97} studies the perception of scene dynamics to interpret image sequences. Their method, unlike ours, requires complete geometric specification of the scene.  A rich set of experiments are  performed by \cite{wu15} on \textit{sliding} motion in the lab settings to estimate object mass and friction coefficients. Our method is not limited to \textit{sliding} and works on a wide range of physical scenarios in various types of scenes. 

\textbf{Action Recognition:} Early prediction of activities has been discussed in \cite{ryoo11,pei11,hoai12,lan14}. Our work is quite different since we estimate long-term motions as opposed to the class of actions. 

\textbf{Human object interaction:} Prediction of human action based on object interactions has been studied in  \cite{koppula13}. Prediction of the behavior of humans based on functional objects in a scene has been explored in \cite{xie13}.  Relative motion of objects in a scene are inferred in \cite{fouhey14}. Our work is related to this line of thought in terms of predicting future events from still images. But our objective is quite different. We do not predict the next action, we care about understanding the underlying physics that justifies future motions in still images.

\textbf{Tracking:} Note that our approach is quite different from tracking \cite{isard98,comaniciu03,collins05} since tracking methods are not destined for single image reasoning. \cite{vondrak08} incorporates simulations to properly model human motion and prevent physically impossible hypotheses during tracking.

\begin{figure*}
\centering
  \includegraphics[width=40pc]{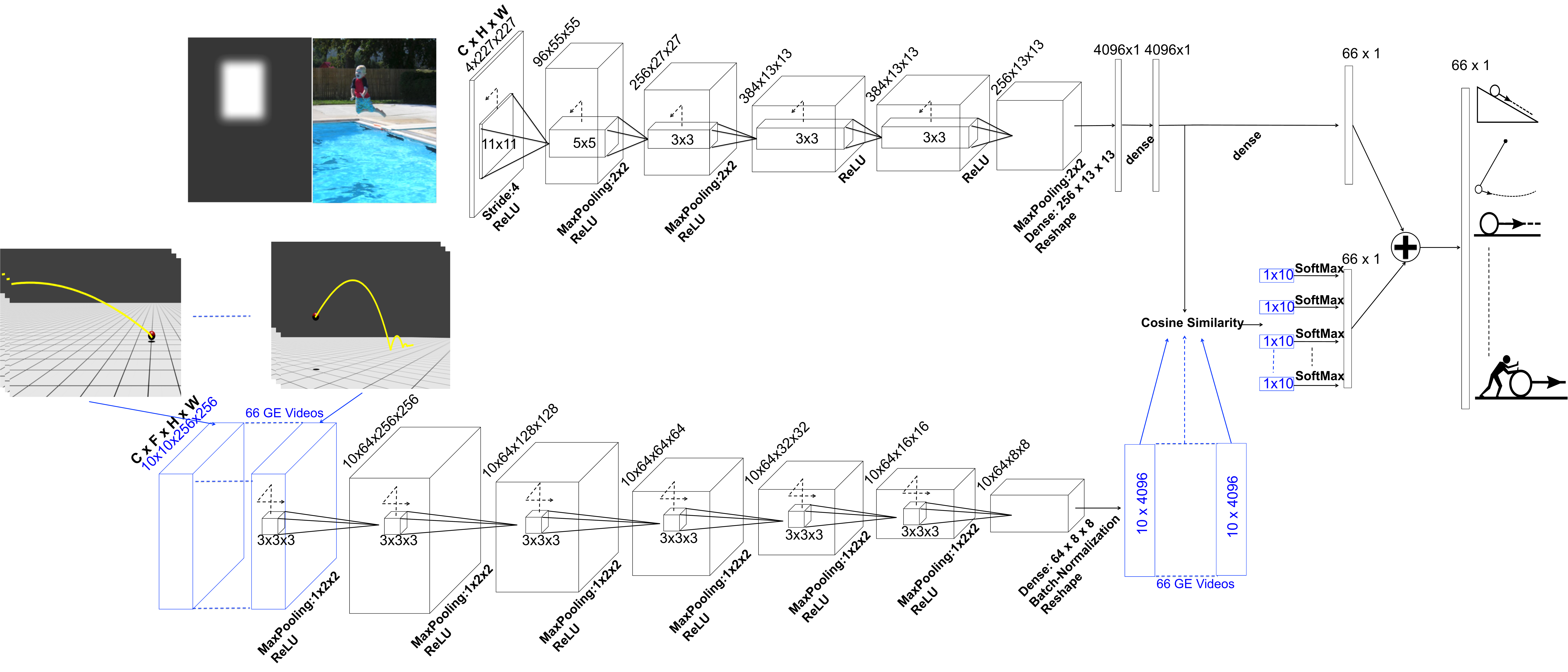}
\caption{ \textbf{Newtonian Neural Network ($N^3$):} This figure illustrates a schematic view of our proposed neural network model. The first row (referred to as \textit{image row}), processes the static image augmented by an extra channel that shows the localization of the query object with a Gaussian-smoothed binary mask. Image row has the same architecture as AlexNet \cite{AlexNet} for image classification. The larger cubes in the row indicate the convolutional outputs. The dimensions for convolutional outputs are \textbf{C}hannels, \textbf{H}eight, \textbf{W}idth. The smaller cubes inside them indicate 2D convolutional filters, which are convolved across Width and Height.  The second row (referred to as \textit{motion row}), processes the video inputs from game engine. This row has similar architecture to C3D \cite{C3D}. The dimensions for convolutional outputs in this row are \textbf{C}hannels, \textbf{F}rames, \textbf{H}eight, \textbf{W}idth. The filters in the motion row are convolved across Frames, Width and Height. These two rows meet by a cosine similarity layer that measures the similarities between the input image and each frame in the game engine videos. The maximum value of these similarities, in each Newtonian scenario is used as the confidence score for that scenario describing the motion of the object in the input image.}
\label{fig:N3}
\end{figure*} 

\section{Problem Statement \& Overview}
Given a static image, our goal is to reason about the expected long-term motion of a query object in 3D. To this end, we use an intermediate physical abstraction called Newtonian scenarios (Figure~\ref{fig:motion}) rendered by a game engine. We learn a mapping from a single image to a state in a Newtonian scenario by our proposed Newtonian Neural Network ($N^3$). A state in a Newtonian scenario corresponds to a specific moment in the video generated by the game engine and includes a set of rich physical quantities (force, velocity, 3D motion) for that moment. Mapping to a state in a Newtonian scenario allows us to borrow the corresponding physical quantities and use them to make predictions about  the long term motion  of the query object in a single image.  

Mapping from a single image to a state in a Newtonian scenario involves solving two problems: (a) figuring out which Newtonian scenario explains the dynamics of the image best; (b) finding the correct moment in the scenario that matches the state of the object in motion. There are strong contextual and visual cues that can help to solve the first problem. However, the second problem involves reasoning about subtle visual cues and is even hard for human annotators.  For example, to predict the expected motion and the current state of the ball in Figure~\ref{fig:teaser} one needs to reason from previous experiences, visual cues, and knowledge about the motion of the object. $N^3$ adopts a data driven approach to use visual cues and the abstract knowledge of motion to learn (a) and (b) at the same time. To encode the visual cues  $N^3$ uses 2D Convolutional Neural Networks (CNN) to represent the image. To learn about motions $N^3$ uses 3D CNNs to represent game engine videos of Newtonian scenarios. By joint embedding $N^3$ learns to map visual cues to exact states in Newtonian scenarios. 
\section{VIND Dataset}
\label{sec:dataset}
We collect VIsual Newtonian Dynamics (VIND) dataset, which contains game engine videos, natural videos and static images corresponding to the Newtonian scenarios. The Newtonian scenarios that we consider are inspired by the way Mechanics is taught in school and cover commonly seen simple motions of objects (Figure~\ref{fig:motion}). Few factors  distinguish these scenarios from each other: (a) the path of the object, \eg scenario (3) describes a projectile motion, while scenario (4) describes a linear motion, (b) whether the applied force is continuous or not, \eg, in scenario (8), the external force is continuously applied, while in scenario (4) the force is applied only in the beginning. (c) whether the object has contact with a support surface or not, \eg, this is the factor that distinguishes scenario (10) from scenario (4).

\noindent \textbf{Newtonian Scenarios:} Representing a Newtonian scenario by a natural video is not ideal due to the noise caused by camera motion, object clutter, irrelevant visual nuisances, \etc.
To abstract away the Newtonian dynamics from noise and clutter in real world, we construct the Newtonian scenarios (shown in Figure~\ref{fig:motion}) using a game engine. A game engine takes a scene configuration as input (\eg a ball above the ground plane) and simulates it forward in time according to laws of motion in physics. For each Newtonian scenario, we render its corresponding game engine scenario from different viewpoints. In total, we obtain 66 game engine videos. For each game engine video, we store its depth map, surface normals and optical flow information in addition to the RGB image. In total each frame in the game engine video has 10 channels.

\noindent \textbf{Images and Videos:} We also collect a dataset of natural videos and images depicting moving objects. The current datasets for action or object recognition are not suitable for our task as they either show complicated movements that go beyond classical dynamics (\eg \textit{head massage} or \textit{make up} in UCF-101 \cite{ucf101}, HMDB-51 \cite{hmdb51}) or  they show no motion (most images in PASCAL \cite{everingham10} or COCO \cite{lin14}). 

\noindent \textbf{Annotations.} We provide three types of annotations for each image/frame: (1) bounding box annotations for the objects that are described by at least one of our Newtonian scenarios, (2) viewpoint information \ie which viewpoint of the game engine videos best describes the direction of the movements in the image/video, (3) state annotations. By state, we mean how far the object has moved on the expected scenario (\eg is it at the beginning of the projectile motion? or is it at the peak point?). More details about the collection of the dataset and the annotation procedure can be found in Section~\ref{sec:experiments}. Example game engine videos corresponding to Newtonian scenario (1) are shown in Figure~\ref{fig:annimages}.

\section{Newtonian Neural Network}
\label{sec:model}
 $N^3$ is shaped by two parallel convolutional neural networks (CNNs); one to encode visual cues and another to represent Newtonian motions. The input to $N^3$ is a static image with four channels (RGBM; where M is the object mask channel that specifies the location of the query object by a bounding-box mask smoothed with a Gaussian kernel) and 66 videos of Newtonian scenarios\footnote{From now on, we refer to the game engine videos rendered for Newtonian scenarios as Newtonian scenarios.}(as described in Section~\ref{sec:dataset}) where each video has 10 frames (equally-spaced frames sampled from the entire video) and each frame has 10 channels (RGB, flow, depth, and surface normal). The output of $N^3$ is a 66 dimensional vector where each dimension shows the confidence of the input image being assigned to a viewpoint of a Newtonian scenario. 
 $N^3$ learns the mapping by enforcing similarities between the vector representations of static images and that of video frames corresponding to Newtonian scenarios. The state prediction is achieved by finding the most similar frame to the static image in the Newtonian space. 

Figure \ref{fig:N3} depicts a schematic illustration of $N^3$. The first row resembles the standard CNN architecture for image classification introduced by~\cite{AlexNet}. We refer to this row as \textit{image row}. Image row has five 2D CONV layers (convolutional layers) and two FC layers (fully connected layers). The second row is a volumetric convolutional neural network inspired by \cite{C3D}. We refer to this row as \textit{motion row}. Motion row has six 3D CONV layers and one FC. The input to the motion row is a batch of 66 videos (corresponding to 66 Newtonian scenarios rendered by game engines). The motion row generates a 4096x10 matrix as output for each video, where a column in this matrix can be seen as a descriptor for a frame in the video. To preserve the same number of frames in the output, we eliminate MaxPooling over the temporal dimension for all CONV layers in the motion row. The two rows are joined by a matching layer that uses cosine similarity as a matching measure. The input to the image row is an RGBM image and the output is a 4096 dimensional vector (values after FC7 layer). This vector can be seen as a visual descriptor for the input image.

The matching layer takes the output of the image row and the output of the motion row as input and computes the cosine similarity between the image descriptors and all of the 10 frames' descriptors in each video in the batch. Therefore, the output of matching layer are 66 vectors where each vector has 10 dimensions. The dimension with maximum similarity value indicates the state of dynamics for each Newtonian scenario. For example, if the third dimension has the maximum value, it means, the input image has maximum similarity with the third frame of the game engine video, thus it must have the same state as that of the third frame in the corresponding game engine video.  SoftMax layers are appended after the cosine similarity layer to pick the maximum similarity as a confidence score for each Newtonian scenario. 
This enables $N^3$ to learn the state prediction without any state level annotations. This is an advantage for $N^3$ that can implicitly learn the state of the motion by directly optimizing for the prediction of Newtonian scenarios. These confidence scores are linearly combined with the confidence scores from the image row to produce the final scores. This linear combination is controlled by a parameter $\lambda \in [0,1]$ that weights the effect of motion for the final score.   

{\bf Training:}  
In order to train $N^3$, we feed the input by picking a batch of random images from the training set and a batch of game engine videos that cover all Newtonian scenarios (66 videos). Each iteration involves a forward and a backward pass through the network. We use negative log-likelihood as our loss function:$E = -\frac{1}{n}\sum_{i=1}^{n}[p_i\log{\hat{p}_i} + (1-p_i)\log{(1-\hat{p_i})]}$, where
$p_i$ is the ground truth probability of the input image being assigned to each Newtonian scenario and $\hat{p}_i$ is the predicted probability obtained by taking SoftMax over the output of $N^3$.  In each iteration, we feed a random batch of images to the network, but a fixed batch of videos across all iterations. This enables $N^3$ to penalize the error over all of the Newtonian scenarios at each iteration. The other option could be passing a pair of a random image and a game engine video, then predicting a binary output showing whether the image corresponds to the Newtonian scenario or not. This requires a lot more iterations to see all the possible positive and negative pairings for an image and has shown to be less effective for our problem.

{\bf Testing:} At test time, the 4096x10 descriptors for abstract motions can be pre-computed from the motion row of $N^3$ after CONV6 layer. For each test, we only feed a single RGBM image as input and obtain the underlying Newtonian scenario $h$ and its matching state $s_h$. The predicted scenario ($h$) is the scenario with maximum confidence in the output. The matching state $s_h$ is achieved by 
\begin{eqnarray}
s_h = \argmax_i\{ Sim(\textbf{x},\textbf{v}_h^i)\}
\end{eqnarray} 
where $\textbf{x}$ is the 4096x1 image descriptor, $\textbf{v}^i_h$ is the 4096x10 video descriptor for Newtonian scenario $h$ and $i\in \{1,2,..,10\}$ indicates the frame index in the video. $Sim(.,.)$ is the standard cosine similarity between two vectors. Given $h$ and $s_h$, a long-term 3D motion path can be drawn for the query object by borrowing the game engine parameters (\eg direction of velocity and force, 3D motion, and  camera view point) from the state $s_h$ of Newtonian scenario $h$.

\vspace{-0.2cm}
\section{Experiments}
\label{sec:experiments}
 We compare our method with a number of baselines in predicting the motion of a query object in an image and provide an ablation study that examines the utility of different components in our method. We further show qualitative results for motion prediction and estimation of force and velocity directions. We also show the benefits of estimating optical flow from our long term motions predicted by our method. Additionally, we show the generalization to unseen scene types.

\begin{figure*}[t]
\centering
  \includegraphics[width=40pc]{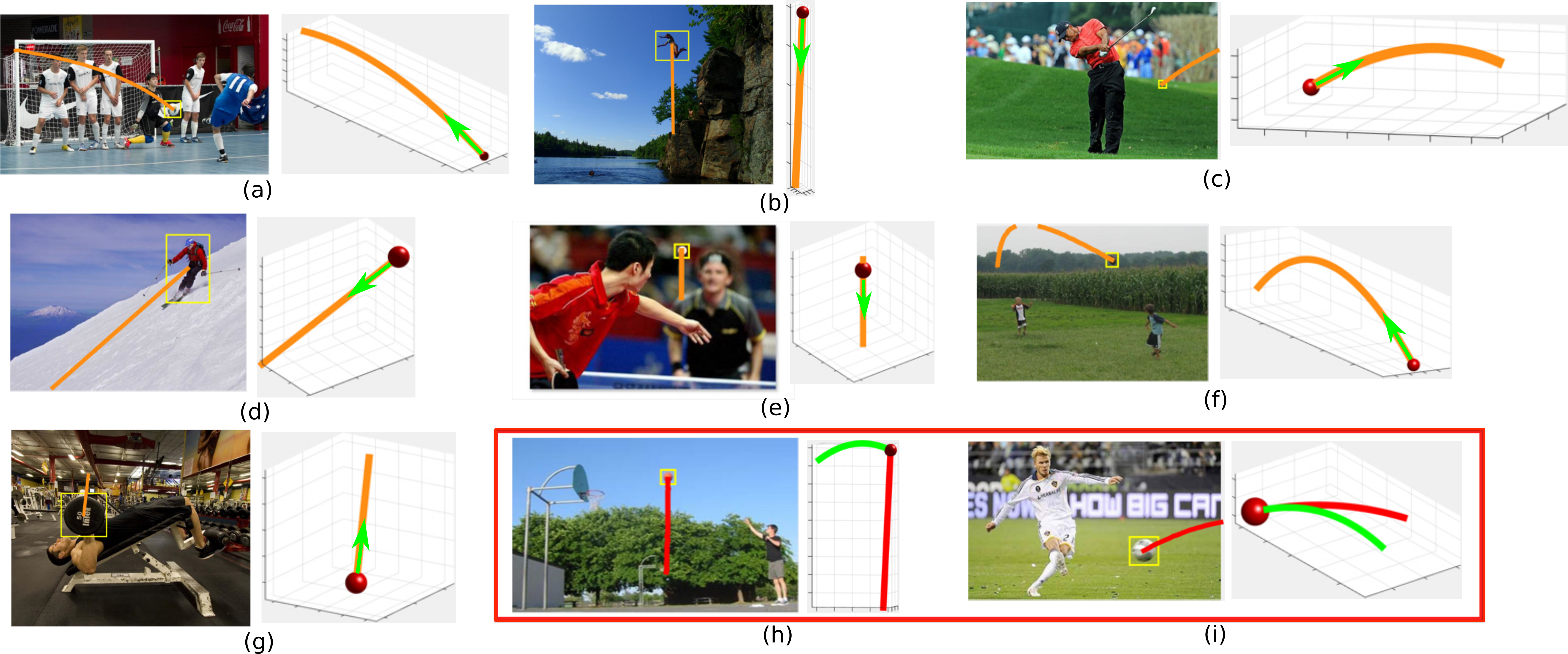}
\caption{The expected motion of the object in the static image is shown in orange. We have visualized the 3D motion of the object (red sphere) and its superposition on the image (left image). We also show failure cases in the red box, where the red and green curves represent our prediction and ground truth, respectively.}
\vspace{-0.3cm}
\label{fig:results}
\end{figure*}

\subsection{Settings}
\label{sec:networksettings}

\textbf{Network:} We implemented our proposed neural network $N^3$ in Torch \cite{torch7}. We use a machine with a 3.5GHz Intel Xeon CPU and GeForce TITAN X GPU to train and test our model. To train $N^3$, we initialized the image row (refer to Figure~\ref{fig:N3}) by a publicly available \footnote{https://github.com/BVLC/caffe/tree/master/models/bvlc\underline{\space}alexnet}  pre-trained CNN model. We initialize the fourth channel (M) by random values drawn from a Gaussian distribution ($\mu=0$,$\sigma=\frac{10}{filter\,size}$). The motion row was initialized randomly, where the random parameters came from a Gaussian distribution ($\mu=0$,$\sigma=\frac{10}{filter\,size}$). For training, we use batches of 128 input images in the image row and 66 videos in the motion row. We run the forward and backward passes for 5000 iterations\footnote{ In our experiments the loss values start converging after 5K iterations.}. We started by the learning rate of $10^{-1}$ and gradually decreased it down to $10^{-4}$.  

In order to prevent the numerical instability of the cosine similarity function, we use the smooth version of cosine similarity, which is defined as: $S(x,y) = \frac{x.y}{|x||y|+\epsilon}$, where $\epsilon = 10^{-5}$.

\textbf{Dataset details:}  We use Blender \cite{blender} game engine to render the game engine videos corresponding to the 12 Newtonian scenarios. We factor out the effect of force magnitude and camera distance. 

The Newtonian scenarios are rendered from 8 different azimuth angles. Scenarios 6, 7, and 11 in Figure~\ref{fig:motion} are symmetric across different azimuth angles and we therefore render them from 3 different elevations of the camera. The Newtonian scenarios 2 and 12 are the same across viewpoints with $180^{\circ}$ azimuth difference. We consider four views for those scenarios. For \textit{stability} (scenario (5)), we consider only 1 viewpoint (there is no motion). In total, we obtain 66 videos for all 12 Newtonian scenarios.

Our new dataset (VIND) contains 6806 video clips in natural scenes. These videos contain 394,807 frames in total. For training, we use frames randomly sampled from these video clips. To train our model, we use bounding box information of query objects and viewpoint annotations for the corresponding Newtonian scenario (the procedure for viewpoint annotations is shown in Figure~\ref{fig:annimages}).

The image portion of our dataset includes 4516 images that are divided into 1458 and 3058 images for validation and testing, respectively. We tune our parameters using the validation set and report our results on the test subset. For evaluation, each image has bounding box, viewpoint and state annotations. 

\begin{table*}[t]
\setlength{\tabcolsep}{1pt}
\centering
\rowcolors{2}{}{gray!35}
\begin{tabular}{l|cccccccccccc|c}
\toprule[0.2 em]%
& {\includegraphics[width = 1.4cm]{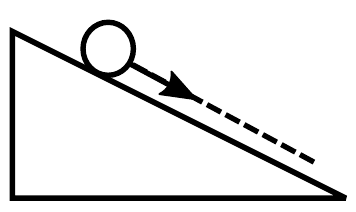}} & {\includegraphics[width = 1.4cm]{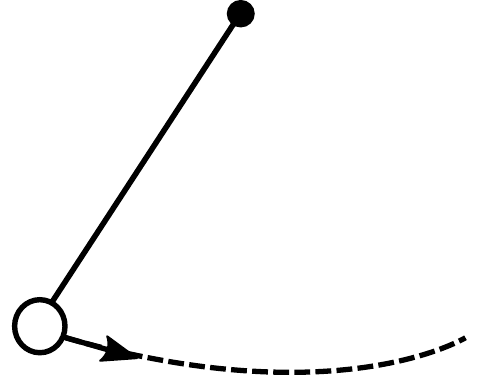}} & {\includegraphics[width = 1.4cm]{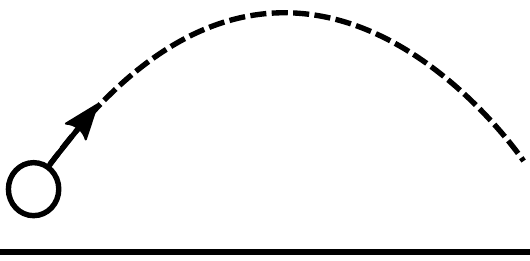}} & {\includegraphics[width = 1cm]{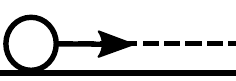}} & {\includegraphics[width = 0.7cm]{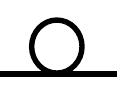}} & {\includegraphics[width = 0.5cm]{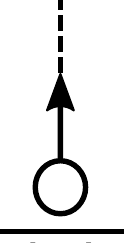}} & {\includegraphics[width = 0.3cm]{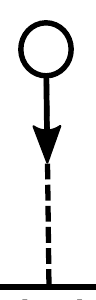}} & {\includegraphics[width = 1.3cm]{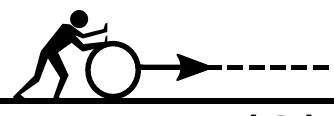}} & {\includegraphics[width = 1.5cm]{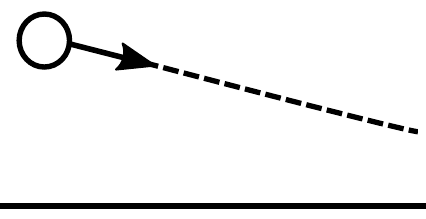}} & {\includegraphics[width = 1.1cm]{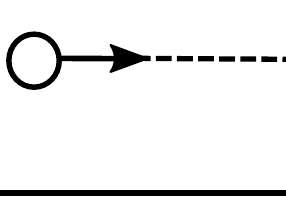}} & {\includegraphics[width = 0.7cm]{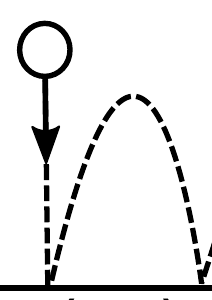}} & {\includegraphics[width = 0.5cm]{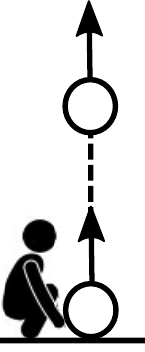}} & Avg.\\
\bottomrule[0.1 em]
\footnotesize{Direct Regression} & 32.7 & 59.9 & 12.4 & 16.1 & 84.6 & 48.8 & 8.2 & 20.2 & 1.6 & 13.8 & 49.0 & 16.4 & 30.31\\
\footnotesize{Direct Regression - Nearest} & 52.7 & 38.4 & 17.3 & 23.5 & 64.9 & \textbf{69.2} & 18.1 & 36.2 & 3.2 & 20.4 & \textbf{76.5} & 24.2 & 37.05\\ 
$N^3$ \footnotesize{(ours)}& \textbf{60.8} & \textbf{64.7} & \textbf{39.4} & \textbf{37.6} & \textbf{95.4} & 54.1 & \textbf{50.3} & \textbf{76.9} & \textbf{9.4} & \textbf{38.1} & 72.1 & \textbf{72.4} & \textbf{55.96}\\
\bottomrule[0.1 em]
\end{tabular}
\caption{Estimation of the motion of the objects in 3D. F-measure is used as the evaluation metric.}
\label{tab:res}
\end{table*}

\subsection{Estimating the motion of query objects}
Given a single image and a query object, we evaluate how well our method can estimate the motion of the object. We compare the resulting 3D curves from our method with that of the ground truth. 

\textbf{Evaluation Metric.} We use an evaluation metric which is similar to the F-measure used for comparing contours (\eg \cite{amfm_pami2011}). The 3D curve of groundtruth and the estimated motion are in $XYZ$ space.  However, the two curves do not necessarily have the same length. 
We slide the shorter curve over the longer curve to find an alignment with the minimum distance. We then compute precision and recall by thresholding the distance between corresponding points on the curves.

We also report results using the Modified Hausdorff Distance (MHD), however the F-measure is more interpretable since it is a number between 0 and 100. 

\textbf{Baselines.}  A set of comparisons with a number of baselines are presented in Table~\ref{tab:res}. The first baseline, called \textit{Direct Regression}, is a direct regression from images to the trajectories in the 3D space (groundtruth curves are represented by B-splines with 1200 knots). For this baseline, we modify AlexNet architecture to regress each image to its corresponding 3D curve. More specifically, we replace the classification loss layer with a Mean Squared Error (MSE) loss layer. Table~\ref{tab:res} shows that $N^3$ significantly outperforms this baseline that aims at directly regressing the motion from visual data. We postulate that this is mainly due to the dimensionality of the output and the complex interplay between subtle visual cues and the 3D motion of objects. To further probe that if the direct regression can even roughly estimate the shape of the trajectory we build an even stronger baseline. For this new baseline, called \textit{Direct Regression-Nearest}, we use the output of the direct regression baseline above to find the most similar 3D curve among Newtonian scenarios (based on normalized Euclidean distance between the B-spline representations). Table~\ref{tab:res} shows that $N^3$ also outperforms this competitive baseline. In terms of the MHD metric, $N^3$ also outperforms the baselines ($5.59$ versus $5.97$ and $7.32$ for the baseline methods; lower is better). 

Figure~\ref{fig:results} shows qualitative results in estimating the expected motion of the object in still images. When $N^3$ predicts a 3D curve for an image it also estimates the viewpoint. This allows us to project the 3D curve back onto the image. Figure~\ref{fig:results} shows examples of these estimated motions. For example, $N^3$ correctly predicts the motion of the football thrown (Figure~\ref{fig:results}(f)), and estimates the right motion for the ping pong ball falling (Figure~\ref{fig:results}(e)).  Note that $N^3$ cannot reason about possible future collisions with other elements in the scene. For example Figure~\ref{fig:results}(a) shows a predicted motion that goes through soccer players. This figure also shows some examples of failures. The mistake in Figure~\ref{fig:results}(h) can be attributed to the large distance between the player and the basketball. Note that when we project 3D curves to images we need to make assumptions about the distance to the camera and the 2D projected curves might have inconsistent scales.

\textbf{Ablation studies.} To study our method in further details, we test two variations of our method. In the first variation, $\lambda$ (defined in Section~\ref{sec:model}) is set to 1, which means that we are ignoring the motion row in the network. We refer to this variation as $N^3 - NV$ in Table~\ref{tab:ablation}. $N^3$ outperforms $N^3 - NV$, indicating that the motion abstraction is an important factor in $N^3$. To study the effectiveness of $N^3$ in state prediction, in the second variation, we measure the utility of providing state supervision for training $N^3$. We modified the output layer of $N^3$ to learn the exact state of the motion from the groundtruth augmented by state level annotations. This case is referred to as $N^3+SS$ in Table~\ref{tab:ablation}. The small gap between the results in $N^3$ and $N^3+SS$ shows that $N^3$ can reliably predict the correct state without state supervision. 

\begin{table}
\centering
\begin{tabular}{|c|c|c|c|}
\hline 
Ablations & $N^3-NV$ & $N^3$ & $N^3+SS$ \\
\hline
F-measure & 52.67 & 55.96 & 56.10 \\
\hline
\end{tabular}
\vspace{2mm}
\caption{Ablation study of 3D motion estimation. The average across 12 Newtonian scenarios is reported.}
\vspace{-5mm}
\label{tab:ablation}
\end{table}

Another ablation is to study the effectiveness of $N^3$ in classifying images into 66 classes corresponding to 12 Newtonian scenarios rendered from different viewpoints. In this ablation, shown in Table~\ref{tab:viewpoint}, we compare $N^3$ to $N^3-NV$ with and without state supervision ($SS$) in a classification setting (not prediction of the motion). Also, our experiments show that $N^3$ and $N^3-NV$ make different types of mistakes since fusing these variations in an optimal way (by an oracle) results in an improvement in classification (25.87).

\begin{figure*}[tp]
\centering
\vspace{-0.4cm}
  \includegraphics[width=35pc]{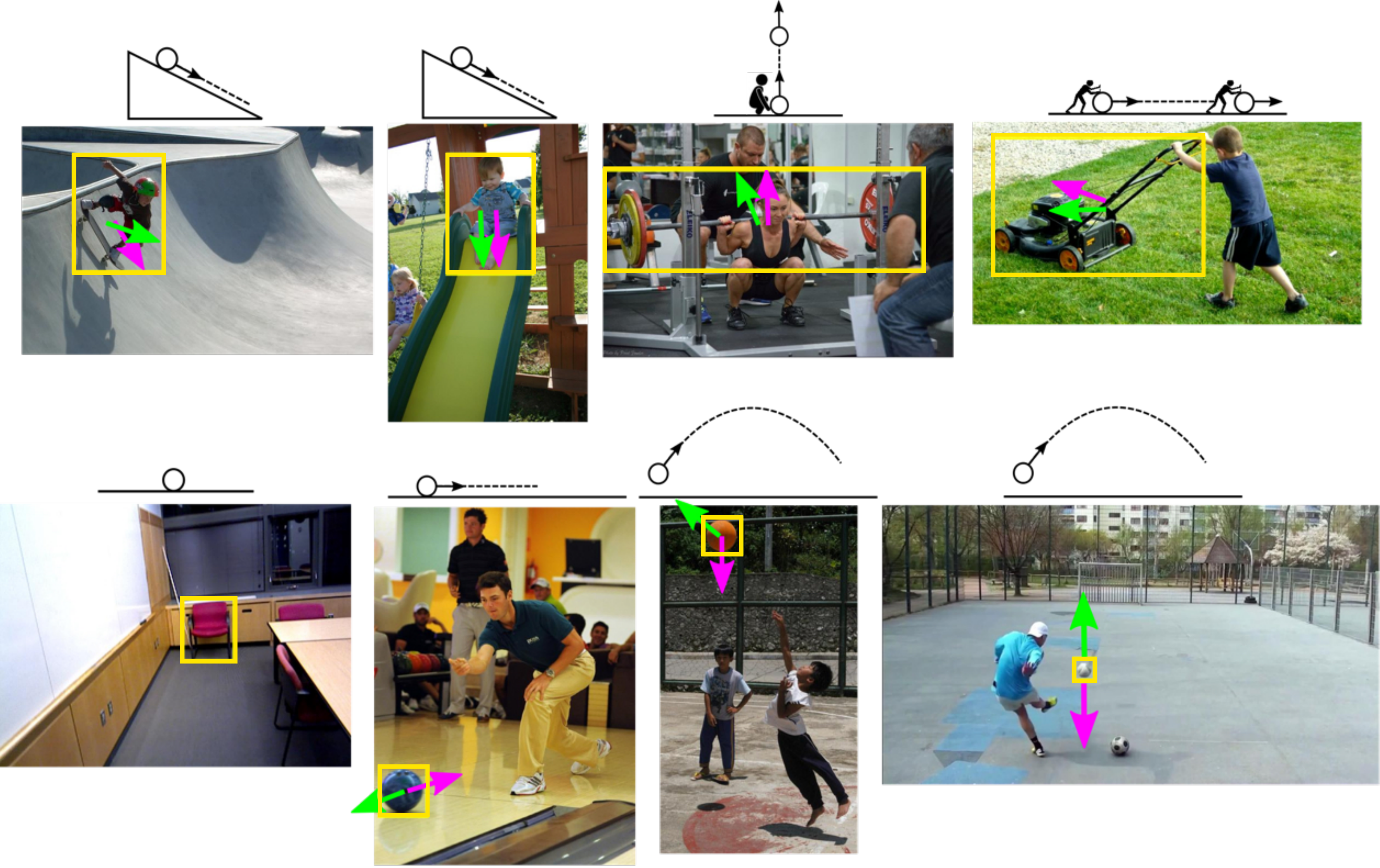}
\caption{Visualization of the \emph{direction} of net force and object velocity. The \textcolor{green}{velocity} is shown in green and the \textcolor{magenta}{net force} is shown in magenta. The corresponding Newtonian scenario is shown above each image. }
\vspace{-0.4cm}
\label{fig:force}
\end{figure*} 

\begin{table} [h]
\setlength{\tabcolsep}{3pt}
\small
\centering
\begin{tabular}{ |c|c|c|c|c| } 
 \hline
 Ablations      & $N^3-NV$  & $N^3-NV+SS$   & $N^3$ & $N^3+SS$\\
 \hline
 Avg. Accuracy  & 20.37     &   19.32       & 21.71 & 21.94\\
 \hline
\end{tabular}
\vspace{1mm}
\caption{Estimation of Newtonian scenario and viewpoint (no state estimation).}
\vspace{-2mm}
\label{tab:viewpoint}
\end{table}

\vspace{-0.2cm}
\textbf{Short-term flow estimation.} Our method is designed to predict long-term motions in 3D, yet it can estimate short term motions by projecting the long term 3D motion onto the image. We compare the effectiveness of $N^3$ in estimating the flow with the state of the art methods explicitly trained to predict short-term flow from a \textit{single image}. In particular, we compare with the recent method of Predictive-CNN \cite{walker15}. For each query object, we average the dense flow predicted by \cite{walker15} over the pixels in the object box and obtain a single flow vector. The evaluation metric is angular error (we do not compute flow magnitude). As shown in Table~\ref{tab:flow}, our method outperforms \cite{walker15} on our dataset. 

\begin{table} [h]
\setlength{\tabcolsep}{3pt}
\centering
\begin{tabular}{ |c|c| } 
 \hline
 Method & Angular Err.\\
 \hline
 Predictive-CNN \cite{walker15} & 1.53 \\
 \hline
 $N^3$ (ours) & \textbf{1.29} \\
 \hline
\end{tabular}
\caption{Short-term flow prediction in a single image. The evaluation metric is angular error.}
\vspace{-0.4cm}
\label{tab:flow}
\end{table}

\textbf{Force and velocity estimation.} It is interesting to see that $N^3$ can predict the \textit{direction} of the net force and velocity in a static image for a query object! Figure~\ref{fig:force} shows qualitative examples. For example, it is exciting to show that $N^3$ can predict the friction in the bolling example, and the gravity in the basketball example. The net force applied to the chair in the bottom row (left) is zero since the normal force from the floor cancels the gravity. 

\textbf{Generalization to unseen scene types.} We also evaluate how well our model generalizes to unseen scene types. We remove all images that represent the same scene type (\eg, all images that show a \textit{billiard} scene in scenario (4)) from our training data and test how well we can estimate the motion of the object in images that show those scene types. Our method outperforms the baseline method (Table~\ref{tab:zeroshot}). The reported result is the average over 12 Newtonian scenarios, where we remove one scene type from each Newtonian scenario during training. 

\begin{table} [t]
\setlength{\tabcolsep}{3pt}
\centering
\begin{tabular}{ |c|c| } 
 \hline
 Method & F-measure \\
 \hline
Direct Regression & 25.76 \\
 $N^3$ (ours) & \textbf{36.40} \\
 \hline
\end{tabular}
\caption{Generalization to unseen scene types.}
\label{tab:zeroshot}
\end{table}

\section{Conclusions}

In this paper we address the challenging problem of Newtonian understanding of objects in static images. Numerous physical quantities contribute to shaping the dynamics of objects in a scene. Direct estimation of those quantities is extremely challenging. In this paper, we assume intermediate physical abstractions, Newtonian scenarios and introduce a model that can map from a single image to a state in a Newtonian scenario. This mapping needs to learn subtle visual and contextual cues to be able to reason about the correct Newtonian scenario, state, viewpoint, etc. Rich physical predictions about the dynamics of objects in an images can then be made by borrowing information through the established correspondences to Newtonian scenarios. This allows us to predict the motion and reason about it in terms of velocity and force directions for a query object in a still image. 

Our current solution can only reason about simple motions of rigid bodies and cannot handle complex and compound motions, specially when it is affected by other external elements in the scene (e.g. the motion of thrown ball would change if there is a wall in front of it in the scene). In addition, our method does not provide estimates for magnitude of the force and velocity vectors. We postulate that there might be very subtle visual cues that can contribute tho those estimates.

Rich physical understanding of images is an important building block towards deeper understanding of images, enables visual reasoning, and opens several new and exciting research directions in scene understanding. Reasoning about how objects move in an image is tightly coupled with semantic and geometric  scene understanding. Explicit joint reasoning about these interactions is an exciting research direction.

{\small
\bibliographystyle{ieee}
\bibliography{egbib}

\begin{thebibliography}{10}\itemsep=-1pt

\bibitem{blender}
{Blender}.
\newblock \url{http://www.blender.org/}.

\bibitem{torch7}
{Torch7}.
\newblock \url{http://torch.ch}.

\bibitem{amfm_pami2011}
P.~Arbelaez, M.~Maire, C.~Fowlkes, and J.~Malik.
\newblock Contour detection and hierarchical image segmentation.
\newblock {\em PAMI}, 2011.

\bibitem{bar09}
M.~Bar.
\newblock The proactive brain: memory for predictions.
\newblock {\em Royal Society of London. Series B, Biological sciences}, 2009.

\bibitem{battaglia13}
P.~Battaglia, J.~Hamrick, and J.~B. Tenenbaum.
\newblock Simulation as an engine of physical scene understanding.
\newblock {\em PNAS}, 2013.

\bibitem{brubaker2008}
M.~A. Brubaker and D.~J. Fleet.
\newblock The kneed walker for human pose tracking.
\newblock In {\em CVPR}, 2008.

\bibitem{brubaker2007}
M.~A. Brubaker, D.~J. Fleet, and A.~Hertzmann.
\newblock Physics-based person tracking using simplified lower-body dynamics.
\newblock In {\em CVPR}, 2007.

\bibitem{brubaker2009}
M.~A. Brubaker, L.~Sigal, and D.~J. Fleet.
\newblock Estimating contact dynamics.
\newblock In {\em ICCV}, 2009.

\bibitem{cheung12}
O.~Cheung and M.~Bar.
\newblock Visual prediction and perceptual expertise.
\newblock {\em Intl. J. of Psychophysiology}, 2012.

\bibitem{collins05}
R.~Collins, Y.~Liu, and M.~Leordeanu.
\newblock On-line selection of discriminative tracking features.
\newblock {\em PAMI}, 2005.

\bibitem{comaniciu03}
D.~Comaniciu, V.~Ramesh, and P.~Meer.
\newblock Kernel-based object tracking.
\newblock {\em PAMI}, 2003.

\bibitem{everingham10}
M.~Everingham, L.~Gool, C.~K. Williams, J.~Winn, and A.~Zisserman.
\newblock The pascal visual object classes (voc) challenge.
\newblock {\em IJCV}, 2010.

\bibitem{fouhey14}
D.~F. Fouhey and C.~Zitnick.
\newblock Predicting object dynamics in scenes.
\newblock In {\em CVPR}, 2014.

\bibitem{hamrick11}
J.~Hamrick, P.~Battaglia, and J.~B. Tenenbaum.
\newblock Internal physics models guide probabilistic judgments about object
  dynamics.
\newblock {\em Annual Meeting of the Cognitive Science Societ}, 2011.

\bibitem{hawkins04}
J.~Hawkins and S.~Blakeslee.
\newblock {\em On Intelligence}.
\newblock Times Books, 2004.

\bibitem{hoai12}
M.~Hoai and F.~{De la Torre}.
\newblock Max-margin early event detectors.
\newblock In {\em CVPR}, 2012.

\bibitem{isard98}
M.~Isard and A.~Blake.
\newblock Condensation conditional density propagation for visual tracking.
\newblock {\em IJCV}, 1998.

\bibitem{jia13}
Z.~Jia, A.~Gallagher, A.~Saxena, and T.~Chen.
\newblock 3d-based reasoning with blocks, support, and stability.
\newblock In {\em CVPR}, 2013.

\bibitem{kitani12}
K.~M. Kitani, B.~D. Ziebart, J.~A.~D. Bagnell, and M.~Hebert.
\newblock Activity forecasting.
\newblock In {\em ECCV}, 2012.

\bibitem{koppula13}
H.~Koppula and A.~Saxena.
\newblock Anticipating human activities using object affordances for reactive
  robotic response.
\newblock In {\em RSS}, 2013.

\bibitem{AlexNet}
A.~Krizhevsky, I.~Sutskever, and G.~E. Hinton.
\newblock Imagenet classification with deep convolutional neural networks.
\newblock In {\em NIPS}, 2012.

\bibitem{hmdb51}
H.~Kuehne, H.~Jhuang, E.~Garrote, T.~Poggio, and T.~Serre.
\newblock Hmdb: a large video database for human motion recognition.
\newblock In {\em ICCV}, 2011.

\bibitem{lan14}
T.~Lan, T.~Chen, and S.~Savarese.
\newblock A hierarchical representation for future action prediction.
\newblock In {\em ECCV}, 2014.

\bibitem{lin14}
T.-Y. Lin, M.~Maire, S.~Belongie, J.~Hays, P.~Perona, D.~Ramanan, P.~Dollár,
  and C.~L. Zitnick.
\newblock Microsoft coco: Common objects in context.
\newblock In {\em ECCV}, 2014.

\bibitem{liu11}
C.~Liu, J.~Yuen, and A.~Torralba.
\newblock Sift flow: Dense correspondence across scenes and its applications.
\newblock {\em PAMI}, 2011.

\bibitem{mann97}
R.~Mann, A.~Jepson, and J.~Siskind.
\newblock The computational perception of scene dynamics.
\newblock {\em CVIU}, 1997.

\bibitem{pei11}
M.~Pei, Y.~Jia, and S.-C. Zhu.
\newblock Parsing video events with goal inference and intent prediction.
\newblock In {\em ICCV}, 2011.

\bibitem{pintea14}
S.~L. Pintea, J.~C. van Gemert, and A.~W.~M. Smeulders.
\newblock D{\'{e}}j{\`{a}} vu: - motion prediction in static images.
\newblock In {\em ECCV}, 2014.

\bibitem{ryoo11}
M.~S. Ryoo.
\newblock Human activity prediction: Early recognition of ongoing activities
  from streaming videos.
\newblock In {\em ICCV}, 2011.

\bibitem{ucf101}
K.~Soomro, A.~R. Zamir, and M.~Shah.
\newblock {UCF101}: A dataset of 101 human action classes from videos in the
  wild.
\newblock Technical Report CRCV-TR-12-01, 2012.

\bibitem{C3D}
D.~Tran, L.~Bourdev, R.~Fergus, L.~Torresani, and M.~Paluri.
\newblock Learning spatiotemporal features with 3d convolutional networks.
\newblock In {\em ICCV}, 2015.

\bibitem{vondrak2008}
M.~Vondrak, L.~Sigal, and O.~C. Jenkins.
\newblock Physical simulation for probabilistic motion tracking.
\newblock In {\em CVPR}, 2008.

\bibitem{vondrak08}
M.~Vondrak, L.~Sigal, and O.~C. Jenkins.
\newblock Physical simulation for probabilistic motion tracking.
\newblock In {\em CVPR}, 2008.

\bibitem{walker14}
J.~Walker, A.~Gupta, and M.~Hebert.
\newblock Patch to the future: Unsupervised visual prediction.
\newblock In {\em CVPR}, 2014.

\bibitem{walker15}
J.~Walker, A.~Gupta, and M.~Hebert.
\newblock Dense optical flow prediction from a static image.
\newblock In {\em ICCV}, 2015.

\bibitem{wu15}
J.~Wu, I.~Yildirim, J.~J. Lim, W.~T. Freeman, and J.~B. Tenenbaum.
\newblock Galileo: Perceiving physical object properties by integrating a
  physics engine with deep learning.
\newblock In {\em NIPS}, 2015.

\bibitem{xie13}
D.~Xie, S.~Todorovic, and S.-C. Zhu.
\newblock Inferring dark matter and dark energy from videos.
\newblock In {\em ICCV}, 2013.

\bibitem{yuen10}
J.~Yuen and A.~Torralba.
\newblock A data-driven approach for event prediction.
\newblock In {\em ECCV}, 2010.

\bibitem{zheng14}
B.~Zheng, Y.~Zhao, J.~C. Yu, K.~Ikeuchi, and S.-C. Zhu.
\newblock Detecting potential falling objects by inferring human action and
  natural disturbance.
\newblock In {\em ICRA}, 2014.

\end{thebibliography}
}

\end{document}